\documentclass[11pt,a4paper]{article}
\usepackage[hyperref]{acl2020}
\usepackage{times}
\usepackage{latexsym}
\usepackage{graphicx}
\usepackage{amsmath}
\usepackage{booktabs}
\usepackage{amsfonts,amssymb}
\usepackage{amssymb}
\usepackage{url}
\usepackage{enumitem}

\usepackage{microtype}

\aclfinalcopy 
\def\aclpaperid{817} 

\title{Document Modeling with Graph Attention Networks for Multi-grained Machine Reading Comprehension}

\author{Bo Zheng\textsuperscript{\rm 1}\thanks{\ \ Work was done while this author was
an intern at Microsoft Research Asia.} , Haoyang Wen\textsuperscript{\rm 1}, Yaobo Liang\textsuperscript{\rm 2}, Nan Duan\textsuperscript{\rm 2}, \\
\textbf{Wanxiang Che\textsuperscript{\rm 1}\thanks{\ \ Email corresponding.} , Daxin Jiang\textsuperscript{\rm 3}, Ming Zhou\textsuperscript{\rm 2}, Ting Liu\textsuperscript{\rm 1}} \\
 \textsuperscript{\rm 1}Harbin Institute of Technology, Harbin, China \\
 \textsuperscript{\rm 2}Microsoft Research Asia, Beijing, China \\
 \textsuperscript{\rm 3}STCA NLP Group, Microsoft, Beijing, China \\
 {\tt \{bzheng,hywen,car,tliu\}@ir.hit.edu.cn} \\
 {\tt \{yalia,nanduan,djiang,mingzhou\}@microsoft.com} 
}

\date{}

\begin{document}
\maketitle
\begin{abstract}
Natural Questions is a new challenging machine reading comprehension benchmark with two-grained answers, which are a long answer (typically a paragraph) and a short answer (one or more entities inside the long answer). 
Despite the effectiveness of existing methods on this benchmark, they treat these two sub-tasks individually during training while ignoring their dependencies.
To address this issue, we present a novel multi-grained machine reading comprehension framework that focuses on modeling documents at their hierarchical nature, which are different levels of granularity: documents, paragraphs, sentences, and tokens.
We utilize graph attention networks to obtain different levels of representations so that they can be learned simultaneously.
The long and short answers can be extracted from paragraph-level representation and token-level representation, respectively.
In this way, we can model the dependencies between the two-grained answers to provide evidence for each other.
We jointly train the two sub-tasks, and our experiments show that our approach significantly outperforms previous systems at both long and short answer criteria. 
\end{abstract}

\section{Introduction}
\begin{figure}[t]
\centering
\includegraphics[scale=0.5]{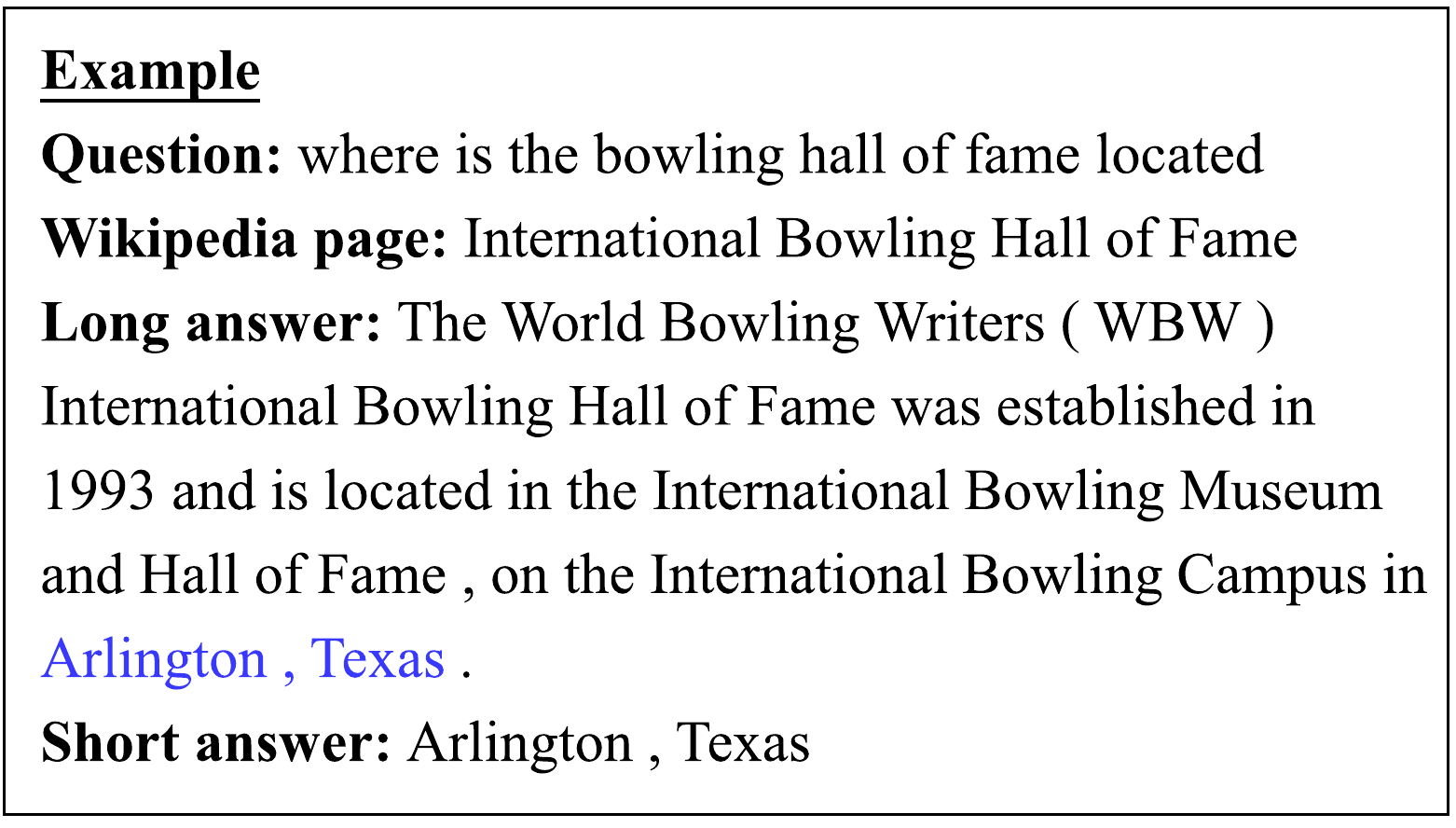}
\caption{An example from NQ dataset.\vspace{-1em}}
\label{fig:example}
\end{figure}

Machine reading comprehension (MRC), a task that aims to answer questions based on a given document, 
has been substantially advanced by recently released datasets and models \citep{rajpurkar2016squad,DBLP:conf/iclr/SeoKFH17,xiong2016dynamic,joshi2017triviaqa,cui2017attention,devlin2018bert,DBLP:conf/acl/GardnerC18}. 
Natural Questions (NQ, \citeauthor{kwiatkowski2019natural}, \citeyear{kwiatkowski2019natural}), a newly released benchmark,
makes it more challenging by introducing much longer documents than existing datasets and questions that are from real user queries.
Besides, unlike conventional MRC tasks (e.g. \citeauthor{rajpurkar2016squad},\citeyear{rajpurkar2016squad}), in NQ, answers are provided in a two-grained format: long answer, which is typically a paragraph, and short answers, which are typically one or more entities inside the long answer. Figure~\ref{fig:example} shows an example from NQ dataset. 

Existing approaches on NQ have obtained promising results. For example, \citet{kwiatkowski2019natural} builds a pipeline model using two separate models: the Decomposable Attention model \cite{DBLP:conf/emnlp/ParikhT0U16} to select a long answer, and the Document Reader model \cite{DBLP:conf/acl/ChenFWB17} to extract the short answer from the selected long answer. Despite the effectiveness of these approaches, they treat the long and short answer extraction as two individual sub-tasks during training and fail to model this multi-grained characteristic of this benchmark, while we argue that the two sub-tasks of NQ should be considered simultaneously to obtain accurate results. 

According to \citet{kwiatkowski2019natural}, a valid long answer must contain all of the information required to answer the question. Besides, an accurate short answer should be helpful to confirm the long answer.  
For instance, when humans try to find the two-grained answers in the given Wikipedia page in Figure~\ref{fig:example}, they will first try to retrieve paragraphs (long answer) describing the entity \textit{bowling hall of fame}, then try to confirm if the \textit{location} (short answer) of the asked entity exists in the paragraph, which helps to finally decide which paragraph is the long answer. 
In this way, the two-grained answers can provide evidence for each other.



To address the two sub-tasks together, instead of using conventional documents modeling methods like hierarchical RNNs \citep{cheng2016neural,yang2016hierarchical,DBLP:conf/aaai/NallapatiZZ17,DBLP:conf/naacl/NarayanCL18}, 
we propose to use graph attention networks \citep{DBLP:conf/iclr/VelickovicCCRLB18} and BERT \cite{devlin2018bert}, 
directly model representations at tokens, sentences, paragraphs, and documents, the four different levels of granularity to capture hierarchical nature of documents. 
In this way, we directly derive scores of long answers from its paragraph-level representations and obtain scores of short answers from the start and end positions on the token-level representations. Thus the long and short answer selection tasks can be trained jointly to promote each other. 
At inference time, we use a pipeline strategy similar to \citet{kwiatkowski2019natural}, where we first select long answers and then extract short answers from the selected long answers.

Experiments on NQ dataset show that our model significantly outperforms previous models at both long and short answer criteria.
We also analyze the benefits of multi-granularity representations derived from the graph module in experiments.

To summarize, the main contributions of this work are as follows:
\begin{itemize}
    \item We propose a multi-grained MRC model based on graph attention networks and BERT.
    \item We apply a joint training strategy where long and short answers can be considered simultaneously, which is beneficial for modeling the dependencies of the two-grained answers.
    \item We achieve state-of-the-art performance on both long and short answer leaderboard of NQ at the time of submission (Jun. 25th, 2019), and our model surpasses single human performance on the development dataset at both long and short answer criteria.
\end{itemize}

We will release our code and models at \url{https://github.com/DancingSoul/NQ_BERT-DM}.

\begin{figure}[t!]
\centering
\includegraphics[trim={0cm 0cm 0cm 0cm},scale=0.65,clip]{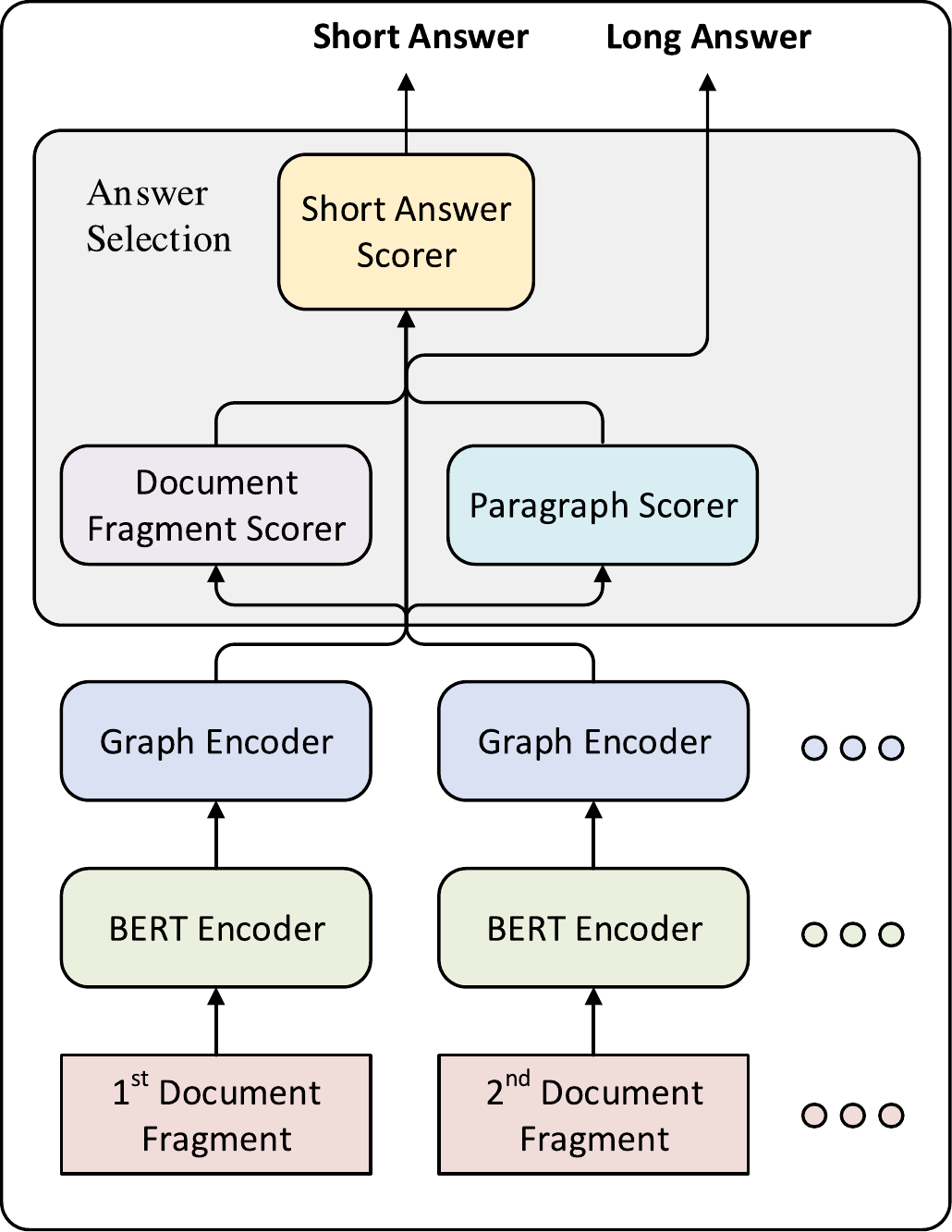}
\caption{System overview. The document fragments of one document are fed into our model independently. The outputs of graph encoders are merged and sent into the answer selection module, which generates a long answer and a short answer.\vspace{-1em}}
\label{fig:system}
\end{figure}

\section{Preliminary}
\subsection{Natural Questions Dataset}
Each example in NQ dataset contains a question together with an entire Wikipedia page. 
The models are expected to predict two types of outputs:
1) long answer, which is an HTML span containing enough information for a reader to completely infer the answer to the question. It can be a paragraph, a table, a list item, or a whole list. A long answer is selected in a list of candidates, or a ``no answer'' should be given if no candidate answers the question;
2) short answer, which can be ``yes'', ``no'' or a list of entities within the long answer. Also, a ``no answer'' should be given if there is no suitable short answer.

\subsection{Data Preprocessing}
\label{dataprocess}
Since the average length of the documents in NQ is too long to be considered as one training instance, we first split each document into a list of document fragments with overlapping windows of tokens, like in the original BERT model for the MRC tasks \citep{alberti2019bert,devlin2018bert}. Then we generate an instance from a document fragment by concatenating a ``[CLS]'' token, tokenized question, a ``[SEP]'' token, tokens from the content of the document fragment and a final ``[SEP]'' token. ``[CLS]'' and ``[SEP]'' follow the definitions from \citet{devlin2018bert}. We tag each document fragment with an answer type as one of the five labels to construct a training instance: ``short'' for instances that contain all annotated short spans, ``yes'' and ``no'' for yes/no annotations where the instances contain the long answer span, ``long'' when the instances contain the long answer span, but there is no short or yes/no answer. In addition to the above situations, we tag a ``no-answer'' to those instances. 

We will explain more details of the data preprocessing in the experiment section.

\section{Approach}

\begin{figure}[t]
\centering
\includegraphics[trim={0cm 1cm 0cm 0cm},scale=0.24,clip]{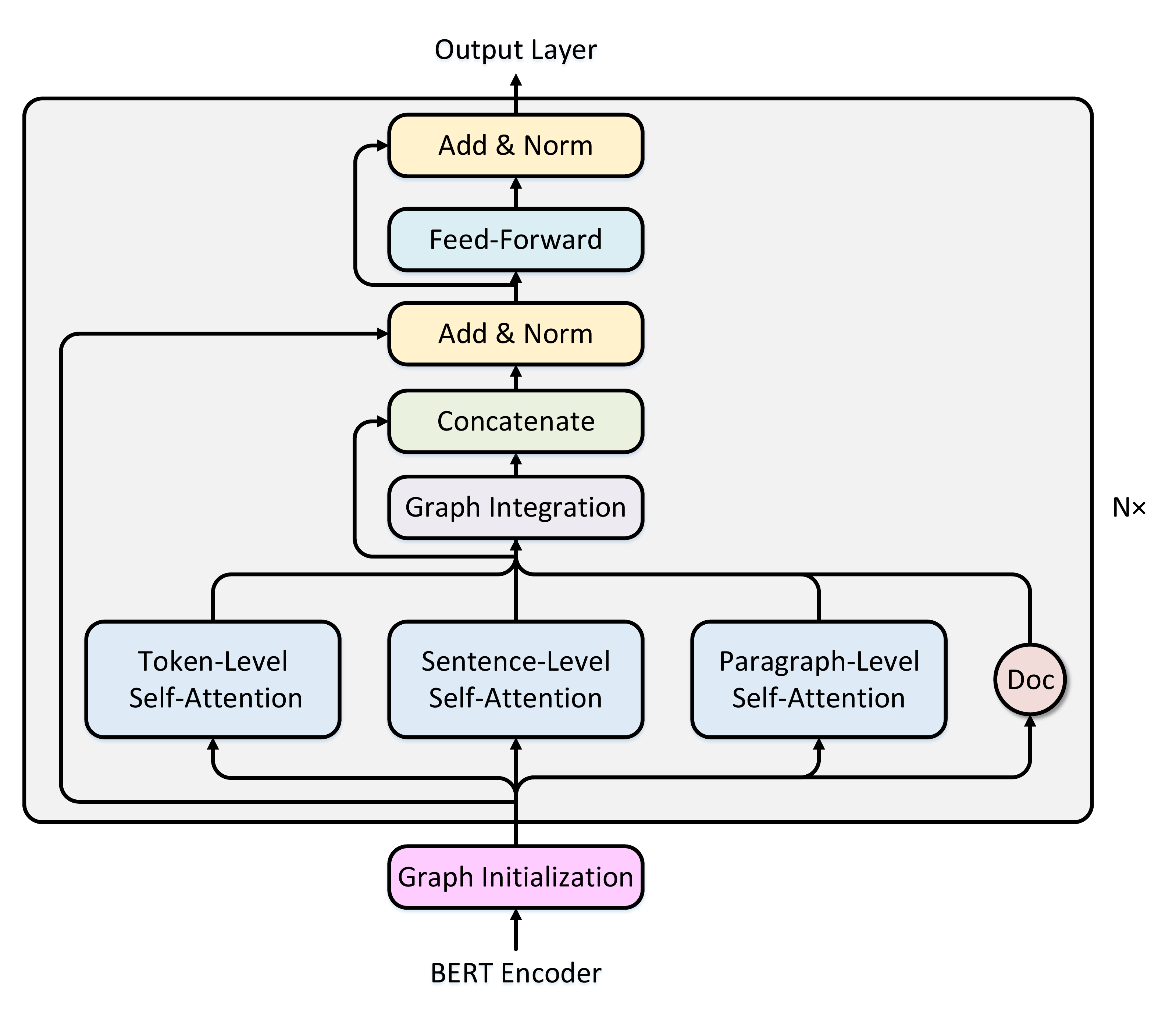}
\caption{Inner structure of our graph encoder.\vspace{-1em}}
\label{fig:modeling}
\end{figure}

In this section, we will explain our model. The main idea of our model lies in multi-granularity document modeling with graph attention networks.
The overall architecture of our model is shown in Figure~\ref{fig:system}. 
\subsection{Input \& Output Definition}

Formally, we define an instance in the training set as a six-tuple
\[
    (\boldsymbol{c}, S, l, s, e, t).
\]

Suppose the instance is generated from the $i$-th document fragment $D_i$ of the corresponding example, then $\boldsymbol{c}=(\text{[CLS]}, Q_{1},...,Q_{|Q|}, \text{[SEP]}, D_{i,1}$ $,...,D_{i,|D_i|}, \text{[SEP]})$ defines the document fragment $D_{i}$ along with a question $Q$ of the instance, $|Q|+|D_{i}|+3=512$ corresponding to the data preprocessing method. 
$S$ denotes the set of long answer candidates inside the document fragment.
$l\in S$ is the target long answer candidate among the candidate set $S$ of this instance. $s$, $e \in \{0, 1, ..., 511\}$ are inclusive indices pointing to the start and end of the target answer span. $t \in \{0, 1, 2, 3, 4\}$ is the annotated answer type, corresponding to the five labels. For instances containing multiple short answers, we set $s$ and $e$ to point to the leftmost position of the first short answer and the rightmost position of the last short answer, respectively. 

Our goal is to learn a model that identifies a long answer candidate $l$ and a short answer span $(s, e)$ in $l$ and predicting their scores for evaluation.

\begin{figure}[t]
\centering
\includegraphics[trim={0cm 0cm 0cm 0cm},scale=0.24, clip]{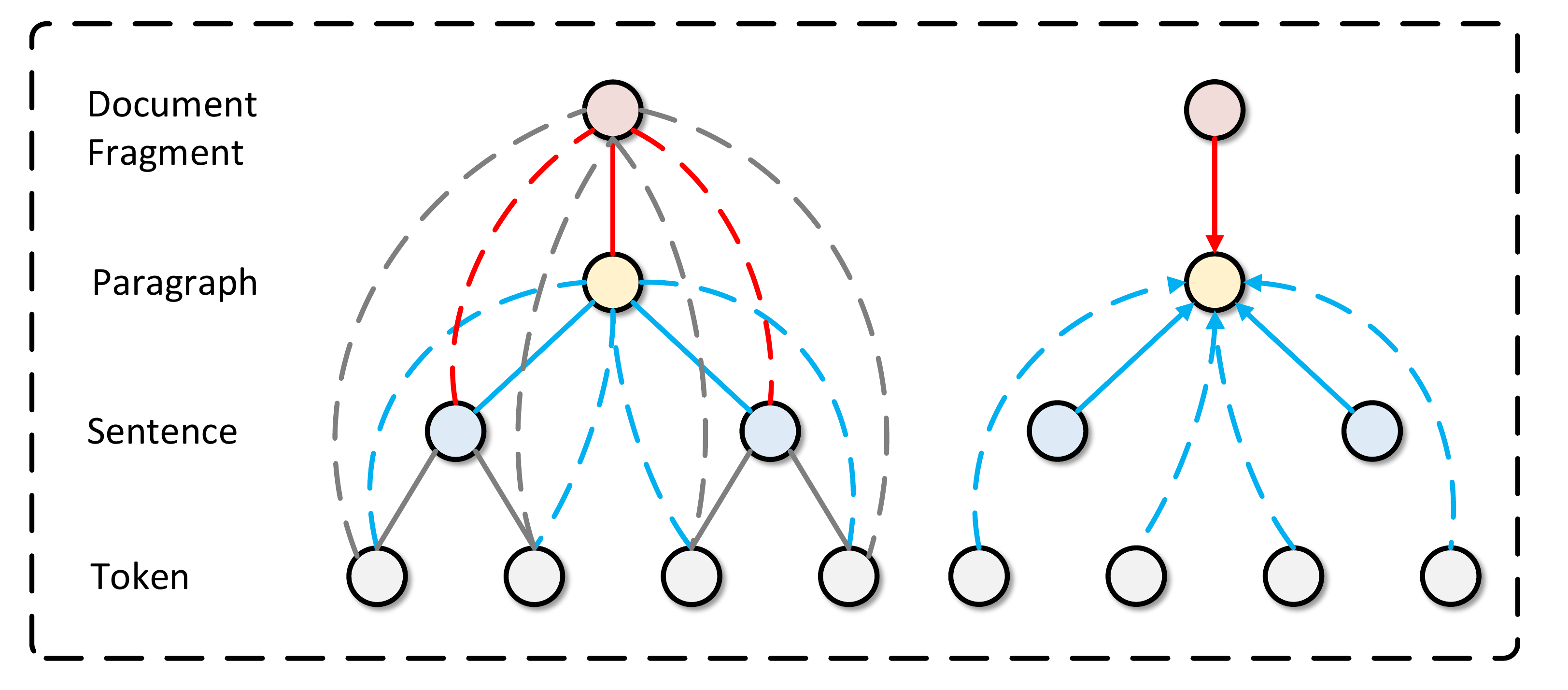}
\caption{The graph on the left is an illustration of the graph integration layer. The graph on the right shows the incoming information when updating a paragraph node. The solid lines represent the edges in the hierarchical tree structure of a document while the dash lines stand for the edges we additionally add.\vspace{-1em}}
\label{fig:integration}
\end{figure}

\subsection{Multi-granularity Document\footnote{For brevity, the word ``document'' refers to document fragment in the rest of our paper.} Modeling}

The intuition of representing documents in multi-granularity is derived from the natural hierarchical structure of a document. 
Generally speaking, a document can be decomposed to a list of paragraphs, which can be further decomposed to lists of sentences and lists of tokens.
Therefore, it is straightforward to treat the document structure as a tree, 
which has four types of nodes, namely token nodes, sentence nodes, paragraph nodes, and a document node.
Different kinds of nodes represent information at different levels of granularity.
Since long answer candidates are paragraphs, tables, or lists,
information at paragraph nodes also represents the information for long answer candidates.

The hierarchical tree structure for a document contains edges that are between tokens and sentences, between sentences and paragraphs, and between paragraphs and documents.
Besides, we further add edges between tokens and paragraphs, between tokens and documents, between sentences and the document to construct a graph. 
All these edges above are bidirectional in our graph representation.
Hence information between every two nodes can be passed through no more than two edges in the graph. 
In the rest of this section, we will present how we utilize this graph structure to pass information between nodes with graph attention networks so that the two-grained answers can promote each other.

\subsection{Graph Encoder}
Figure~\ref{fig:modeling} shows the inner structure of our graph encoder. 
Each layer in our graph encoder consists of three self-attention layers, a graph integration layer, and a feed-forward layer. The self-attention layers are used for interactions among nodes with the same granularity, while the graph integration layer aims at gathering information from other levels of granularity with graph attention networks. Figure~\ref{fig:integration} is an illustration for the graph integration layer. Since self-attention is a special case of graph attention networks, where the graph is fully connected, we only introduce the general form of graph attention networks, which can be generalized to the self-attention mechanism.

\subsubsection{Graph Attention Networks}
We apply graph attention networks \citep{DBLP:conf/iclr/VelickovicCCRLB18} to model the information flow between nodes, which can further improve the representations of nodes by attention mechanism over features from its neighbors. In this way, the interaction between the two-grained answers can be enhanced. Instead of other graph-based models, we use graph attention networks to keep consistency with the multi-head attention module in the BERT model.
We will describe a single layer of our graph attention networks in the following.

We define a graph $\mathcal{G} = \left(\mathcal{V}, \mathcal{E}, X\right)$ that is composed of a set of nodes $\mathcal{V}$, node features $X=(\boldsymbol{h}_{1},...,\boldsymbol{h}_{|\mathcal{V}|})$ and a list of directed edge set $\mathcal{E}=\left(\mathcal{E}_{1},...,\mathcal{E}_{K}\right)$ where $K$ is the number of edges. Each $i \in \mathcal{V}$ has its own representation $\boldsymbol{h}_{i} \in \mathbb{R}^{d_{\text{h}}}$ where $d_{\text{h}}$ is the hidden size of our model.

We use the multi-head attention mechanism in our graph attention networks following \citet{DBLP:conf/nips/VaswaniSPUJGKP17}. We describe one of the $m$ attention heads. All the parameters are unique to each attention head and layer. If there is an edge from node $j$ to node $i$, the attention coefficient $e_{ij}$ is calculated as follows:
\begin{equation}
\label{eij}
    e_{ij}=\frac{\left ({\boldsymbol{h}}_{i} \mathbf{W}^{\text{Q}}\right) \left({\boldsymbol{h}}_{j} \mathbf{W}^{\text{K}}\right)^\textrm{T}}{\sqrt{d_{\text{z}}}}.
\end{equation}

We normalize the attention coefficients of node $i$ by using the \textit{softmax} function across all the neighbor nodes $j\in\mathcal{N}_i$. Especially, there is a self-loop for each node (i.e. $i \in \mathcal{N}_i$) to allow it update itself.
This process can be expressed as:
\[
    \alpha_{ij}=\textrm{softmax}_{j}(e_{ij})=
    \frac{\exp(e_{ij})}{\sum_{k\in\mathcal{N}_i}\exp(e_{ik})}.
\]

Then the output of this attention head $\boldsymbol{z}_i$ is computed as a weighted sum of linear transformed input elements:
\begin{equation}
\label{zi}
    \boldsymbol{z}_{i} = \sum_{j\in\mathcal{N}_i} \alpha_{ij} {\boldsymbol{h}}_{j} \mathbf{W}^{\text{V}}.
\end{equation}

In the above equations, $\mathbf{W}^{\text{Q}}$,$\mathbf{W}^{\text{K}}$ and $\mathbf{W}^{\text{V}} \in \mathbb{R}^{d_\text{h} \times d_\text{z}}$ are parameter matrices, $d_\text{z}$ is the output size of one attention head, we use $d_\text{z} \times m = d_\text{h}$.

Finally we get the multi-head attention result $z_{i}' \in \mathbb{R}^{d_\text{h}}$ by concatenating the outputs of $m$ individual attention heads:
\[
    \boldsymbol{z}_{i}' = \underset{k=1}{\overset{m}{\Arrowvert}}  \boldsymbol{z}_{i}^{k}.
\]

\subsubsection{Self-Attention Layer}
The self-attention mechanism is equivalent to the fully-connected version of graph attention networks. To make interactions among nodes with the same granularity, we utilize three self-attention layers, which are token-level self-attention, sentence-level self-attention, and paragraph-level self-attention. 
Since the four types of nodes are essentially heterogeneous, 
we separate the self-attention layer from the graph integration layer to distinguish information from nodes with the same granularity or different ones.
\subsubsection{Graph Integration Layer}
We use graph attention networks on the graph presented in Figure~\ref{fig:integration}, this layer allows information to be passed to nodes with different levels of granularity. 
Instead of integrating information only once after the graph encoder, we put this layer right after every self-attention layer inside the graph encoder,  which means the update brought by the self-attention layer will also be utilized by the nodes with other levels of granularity. This layer helps to model the dependencies of the two-grained answers.
We concatenate the input and output of the graph integration layer and pass it to the feed-forward layer.
\subsubsection{Feed-Forward Layer}
Following the inner structure of the transformer \cite{DBLP:conf/nips/VaswaniSPUJGKP17}, we also utilize an additional fully connected feed-forward network at the end of our graph encoder. It consists of two linear transformations with a \textsc{gelu} activation in between. \textsc{gelu} is Gaussian Error Linear Unit activation \cite{DBLP:journals/corr/HendrycksG16}, and we use \textsc{gelu} as the non-linear activation, which is consistent with BERT.
\subsubsection{Relational Embedding}
Inspired by positional encoding in \citet{DBLP:conf/nips/VaswaniSPUJGKP17} and relative position representations in \citet{DBLP:conf/naacl/ShawUV18}, we introduce a novel relational embedding on our constructed graph, which aims at modeling the relative position information between nodes on the multi-granularity document structure. We make the edges in our document modeling graph to embed relative positional information. 
We modify equation \ref{eij} and \ref{zi} for $e_{ij}$ and $\boldsymbol{z}_{i}$ to introduce our relational embedding as follows:
\begin{eqnarray*}
    e_{ij}&=&\frac{\left({\boldsymbol{h}}_{i} \mathbf{W}^{\text{Q}}\right) {\left({\boldsymbol{h}}_{j} \mathbf{W}^{\text{K}}\right)}^{\mathrm{T}} + {\boldsymbol{h}}_{i} \mathbf{W}^{\text{Q}}\left(\boldsymbol{a}_{ij}^{\text{K}}\right)^{\mathrm{T}}}{\sqrt{d_{\text{z}}}},\\
    \boldsymbol{z}_{i} &=& \sum_{j\in\mathcal{N}_i} \alpha_{ij}\left(\boldsymbol{h}_{j}\mathbf{W}^{\text{V}} + \boldsymbol{a}_{ij}^{\text{V}}\right).
\end{eqnarray*}

In above equations, the edge between node $i$ and node $j$ is represented by learnable embedding $\boldsymbol{a}_{ij}^K$, $\boldsymbol{a}_{ij}^V \in \mathbb{R}^{d_z}$. The representation can be shared across attention heads. 
Compared to previous work which encodes positional information in the embedding layer, our proposed relational embedding is more flexible, and the positional information can be taken into consideration in each graph layer. For example, relational embedding between two nodes of the same type represents the relative distance between them in the self-attention layer. In the graph integration layer, relational embedding between a sentence and its paragraph represents the relative position of the sentence in the paragraph, and it is the same for other types of edges.
\subsubsection{Graph Initialization}
Since the BERT model can only provide token-level representation, we use a bottom-up average-pooling strategy to initialize the nodes other than token-level nodes. We use $o_{i} \in \{0, 1, 2, 3\}$ to indicate the type of node $i$, representing token node, sentence node, paragraph node and document node respectively. The initialized representation is calculated as follows:
\[
    \boldsymbol{h}_{i}^{0} = \mathop{\text{average}}\limits_{j\in\mathcal{N_\textrm{i}}, o_{j}+1=o_i} \left\{ \boldsymbol{h}_{j}^{0} + {\boldsymbol{a}_{ij}} \right\} + \boldsymbol{b}_{o_i},
\]
where ${\boldsymbol{a}_{ij}}$, $\boldsymbol{b}_{o_i} \in \mathbb{R}^{d_h}$ represent the relational embedding and node type embedding in the graph initializer.

\begin{table*}[ht]
\centering
\footnotesize
\begin{tabular}{lccccccccccccc}
\toprule
 & \multicolumn{3}{c}{Long Answer Dev} & \multicolumn{3}{c}{Long Answer Test} & & \multicolumn{3}{c}{Short Answer Dev} & \multicolumn{3}{c}{Short Answer Test} \\
 & P & R & F1 & P & R & F1 & & P & R & F1 & P & R & F1 \\ \midrule
DocumentQA & 47.5 & 44.7 & 46.1 & 48.9 & 43.3 & 45.7 & & 38.6 & 33.2 & 35.7 & 40.6 & 31.0 & 35.1 \\
DecAtt + DocReader & 52.7 & 57.0 & 54.8 & 54.3 & 55.7 & 55.0 & & 34.3 & 28.9 & 31.4 & 31.9 & 31.1 & 31.5 \\ 
$\textrm{BERT}_{\textrm{joint}}$ & 61.3 & 68.4 & 64.7 & 64.1 & 68.3 & 66.2 & & 59.5 & 47.3 & 52.7 & \textbf{63.8} & 44.0 & 52.1 \\ 
 + 4M synthetic data & 62.3 & 70.0 & 65.9 & 65.2 & 68.4 & 66.8 & & 60.7 & 50.4 & 55.1 & 62.1 & 47.7 & 53.9 \\  \midrule
\textbf{BERT-syn+Model-III} & 72.4 & 73.0 & 72.7 & - & - & - & & 60.1 & 54.1 & 56.9 & - & - & - \\
\textbf{+ ensemble 3 models} & \textbf{74.2} & \textbf{73.6} & \textbf{73.9} & \textbf{73.7} & \textbf{75.3} & \textbf{74.5} & & \textbf{64.0} & \textbf{54.9} & \textbf{59.1} & 62.6 & \textbf{55.3} & \textbf{58.7} \\ \midrule
Single Human & 80.4 & 67.6 & 73.4 & - & - & - & & 63.4 & 52.6 & 57.5 & - & - & - \\ 
Super-annotator & 90.0 & 84.6 & 87.2 & - & - & - & & 79.1 & 72.6 & 75.7 & - & - & - \\ \bottomrule
\end{tabular}
\caption{Results of our best model on NQ compared to the previous systems and to the performance of a single human annotator and of an ensemble of human annotators. The previous systems include DocumentQA \cite{DBLP:conf/acl/GardnerC18}, DecAtt + DocReader \cite{DBLP:conf/emnlp/ParikhT0U16,DBLP:conf/acl/ChenFWB17} , $\textrm{BERT}_{\textrm{joint}}$ and $\textrm{BERT}_{\textrm{joint}}$ + 4M synthetic data \cite{DBLP:conf/acl/AlbertiAPDC19}.}
\label{table:nq-results}
\end{table*}

\subsection{Output Layer}
The objective function is defined as the negative sum of the log probabilities of the predicted distributions, averaged over all the training instances. The log probabilities of predicted distributions are indexed by the true start and end indices, true long answer candidate index, and the type of this instance. 
\begin{equation*}
\begin{aligned}
L(\theta) = &-\frac{1}{N}\sum_{i}^{N}~[ \log p(s, e, t, l\mid \mathbf{c}, S)] \\
 = &-\frac{1}{N}\sum_{i}^{N}~[ \log p_{s}(s\mid\mathbf{c},S) + \log p_{e}(e\mid\mathbf{c},S) \\
 &+ \log p_{t}(t\mid\mathbf{c}, S) + \log p_{l}(l\mid\mathbf{c},S)],
\end{aligned}
\end{equation*}
where $p_{s}(s\mid\mathbf{c}, S)$, $p_{e}(e\mid\mathbf{c}, S)$, $p_{l}(l\mid\mathbf{c}, S)$ and $p_{t}(t\mid\mathbf{c}, S)$ are the probabilities for the start and end position of the short answer, probabilities for the long answer candidate, and probabilities for the answer type of this instance, respectively. One of the probability, $p_{s}(s\mid\mathbf{c}, S)$, is computed as follow, and the others are similar to it:
\label{outputlayer}
\[
    p_{s}(s\mid\mathbf{c}, S) = \textrm{softmax}(f_{s}(s,\mathbf{c}, S;\theta)),
\]
where $f_{s}$ is a scoring function, derived from the last layer of graph encoder. Similarly, we derive score functions at the other three levels of granularity.
For instances without short answers, we set the target start and end indices to the ``[CLS]'' token.
We also make ``[CLS]'' markup as the first sentence and paragraph, and the paragraph-level ``[CLS]'' will be classified as long answers for the instances without long answers.
At inference time, we get the score of a document fragment $g(\mathbf{c}, S)$, long answer score $g(\mathbf{c}, S, l)$ and short answer score $g(\mathbf{c}, S, s, e)$ as follows:
\begin{equation*}
\begin{aligned}
    g(\mathbf{c}, S) =& f_{t}(t>0, \mathbf{c}, S;\theta) - f_{t}(t=0, \mathbf{c}, S; \theta); \\
    g(\mathbf{c}, S, l) =& f_{l}(l, \mathbf{c}, S; \theta) - f_{l}(l =\text{[CLS]}, \mathbf{c}, S;\theta); \\
    g(\mathbf{c}, S, s, e) =& f_{s}(s, \mathbf{c}, S;\theta) + f_{e}(e, \mathbf{c}, s;\theta) \\
    &- f_{s}(s=\text{[CLS]}, \mathbf{c}, S;\theta) \\
    &- f_{e}(e=\text{[CLS]}, \mathbf{c}, S;\theta).
\end{aligned}
\end{equation*}

We use the sum of $g(\mathbf{c}, S, l)$ and $g(\mathbf{c}, S)$ to select a long answer candidate with highest score. $g(\mathbf{c}, S)$ is considered as a bias term for document fragments. Then we use $g(\mathbf{c}, S, s, e)$ to select the final short answer within the selected long answer span. We rely on the official NQ evaluation script to set thresholds to separate the predictions to positive and negative on both long and short answer. 

\label{sec:experiment}
\section{Experiments}
In this section, we will first describe the data preprocessing details, then give the experimental results and analysis. We also conduct an error analysis and two case studies in the appendix.
\subsection{Data Preprocessing Details}
We ignore all the HTML tags as well as tokens not belonging to any long answer candidates. 
The average length of documents is approximately $4,500$ tokens after this process.
Following \citet{devlin2018bert} and \citet{alberti2019bert}, 
we first tokenize questions and documents using a $30,522$ wordpiece vocabulary. 
Then we slide a window of a certain length over the entire length of the document with a stride of $128$ tokens, generating a list of document fragments. There are about 7 paragraphs and 18 sentences on average per document fragment.
We add special markup tokens at the beginning of each long answer candidate according to the content of the candidate. The special tokens we introduced are of the form ``[Paragraph=N]'', ``[Table=N]'' and ``[List=N]''. According to \citet{alberti2019bert}, this decision was based on the observation that the first few paragraphs and tables in the document are more likely to contain the annotated answer.
We generate $30$ instances on average per NQ example, and each instance will be processed independently during the training phase.

Since the fact that only a small fraction of generated instances are tagged as positive instances which contains a complete span of long or short answer, and that $51\%$ of the documents do not contain the answers for the questions,
We downsample about $97\%$ of null instances to get about $660,000$ training instances in which $350,000$ has a long answer, and $270,000$ has short answers.


\subsection{Experimental Settings}
We use three model settings for our experiments, which are: 1) Model-I: A refined BERT baseline on the basis of \citet{alberti2019bert}; 2) Model-II: A pipeline model with only graph initialization method to get representation of sentence, paragraph, and document; 3) Model-III: Adding two layers of our graph encoder on the basis of Model-II.

Model-I improves the baseline in \citet{alberti2019bert} in two ways: 
1) When training an instance with a long answer only, we ignore the loss of predicting the short answer span to ``no-answer'' because it would introduce distraction to the model. 
2) We sample more negative instances.

We use three BERT encoders to initialize our token node representation: 1) BERT-base: a BERT-base-uncased model finetuned on SQuAD 2.0; 2) BERT-large: a BERT-large-uncased model finetuned on SQuAD 2.0; 3) BERT-syn: Google's BERT-large-uncased model pre-trained on SQuAD2.0 with N-Gram Masking and Synthetic Self-Training.\footnote{This model can be downloaded at \url{https://bit.ly/2w7nUQK}.} 
Since the Natural Question dataset does not provide sentence-level information, we additionally use spacy \cite{spacy2} as the sentence segmentor to get the boundaries of sentences.

We trained the model by minimizing loss $L$ from Section~\ref{outputlayer} using the Adam optimizer \cite{DBLP:journals/corr/KingmaB14} with a batch size of $32$. We trained our model for $2$ epochs with an initial learning rate of $2\times 10^{-5}$, and we use a warmup proportion of $0.1$. The training of our proposed model is conducted on $4$ Tesla P40 GPUs for approximately 2 days. For each setting, the results are averaged over three models initialized with different random seeds to get a more solid comparison, which also suggests the improvements brought by our methods are relatively stable. The hidden size, the number of attention heads, and the dropout rate in our graph encoder are equal to the values in the corresponding BERT model.

\begin{table}[t]
\centering
\begin{tabular}{lll}
\toprule
Model & LA. F1 & SA. F1 \\ \midrule
BERT-base+Model-I & 63.9 & 51.0 \\ 
BERT-base+Model-II & 67.7 & 50.9 \\
BERT-base+Model-III & \textbf{68.9} & \textbf{51.9}\\ \midrule
$\textrm{BERT}_{\textrm{joint}}$ & 64.7 & 52.7 \\ 
BERT-large+Model-I & 66.0 & 52.9 \\ 
BERT-large+Model-II & 70.3 & 53.2 \\
BERT-large+Model-III & \textbf{70.7} & \textbf{53.8} \\ \midrule
BERT-syn+Model-I & 67.8 & 56.1 \\ 
BERT-syn+Model-II & 72.2 & 56.7 \\
BERT-syn+Model-III & \textbf{72.7} & \textbf{56.9} \\ \bottomrule
\end{tabular}
\caption{Comparison of different models with different BERT models on the development dataset.}
\label{table:nq-dev}
\end{table}

\subsection{Comparison}
The main results are shown in Table~\ref{table:nq-results}. The results show that our best model BERT-syn+Model-III(ensemble 3 models) have gained improvement over the previous models by a large margin. Our ensemble strategy is to train three models with different random seeds. The scores of answer candidates are averaged over these three models.
At the time of submission (Jun. 25th, 2019), this model has achieved the state-of-the-art performance on both long answer (F1 score of $74.5\%$) and short answer (F1 score of $58.7\%$) on the public leaderboard\footnote{Since we can only make 10 submissions on the test dataset, we only submit and report the result of our best model. Due to the official attempts on the test dataset are given 24 hours. We can only ensemble 3 models at most.}. Furthermore, our model surpasses single human performance at both long and short answer criteria on the development dataset.

The comparison of different models with different BERT models is illustrated in Table~\ref{table:nq-dev}. 
The results show that our approach significantly outperforms our baseline model on both the long answer and the short answer. For the BERT-base setting, our Model-II with a pipeline inference strategy outperforms our baseline by $3.8\%$ on long answer F1 score while our Model-II with two graph layers further improves the performance by $1.2\%$ and $1.0\%$. For the BERT-syn setting, the Model-III benefits less from the graph layers because the pretraining for this model is already quite strong. Our Model-III with BERT-large, compared to previously public model ($\textrm{BERT}_{\textrm{joint}}$) also using BERT-large,  improves long answer F1 score by 6.0\% and short answer F1 score by 1.1\% on the development set. 

From Table~\ref{table:nq-results} and Table~\ref{table:nq-dev}, we can see that the ensemble of human annotators can lead to a massive improvement at both long and short answer criteria (from 73.4\% to 87.2\%, 57.5\% to 75.7\%). However, the improvement of ensembling our BERT-based model is relatively smaller (from 72.7\% to 73.9\%, 56.9\% to 59.1\%). This suggests that the diversity of human annotators is a lot better than the same model structure with different random seeds. How to improve the diversity of the deep learning models for the open-domain datasets like NQ remains as a hard question.




\begin{table}[t]
\centering
\begin{tabular}{lll}
\toprule
 Model & LA.F1 & SA.F1 \\
\midrule
0-layer & 67.7 & 50.9 \\
1-layer & 68.8 & 51.2 \\
2-layer & \textbf{68.9} & \textbf{51.9} \\
3-layer & \textbf{68.9} & \textbf{51.9} \\
4-layer & \textbf{68.9} & 51.7 \\\bottomrule
\end{tabular}
\caption{Influences of graph layer numbers on the development set.}
\label{table:nq-nlayer}
\end{table}

\begin{table}[t]
\centering
\begin{tabular}{lll}
\toprule
Model & LA. F1 & SA. F1 \\ \midrule
BERT-base+Model-III & \textbf{68.9} & \textbf{51.9} \\ \midrule
-Graph module & 63.9 & 51.0 \\
-Long answer prediction & 65.1 & 51.4 \\
-Short answer prediction & 68.2 &  - \\ 
-Relational embedding & 68.8 & 51.7 \\ 
-Graph integration layer & 68.3 & 51.1 \\ 
-Self-attention layer & 68.4 & 51.2 \\ \bottomrule
\end{tabular}
\caption{Ablation study on the development set.}
\label{table:nq-ablation}
\end{table}

\subsection{Ablation Study}
We evaluate the influence of layer numbers, which is illustrated in Table~\ref{table:nq-nlayer}. We can see the increase in the performance of our models when the number of layers increases from $0$ to $2$ (The 0-layer setting means that only the graph initialization module is used to obtain the graph representations). Then the model performance does not improve with the number of network layers increasing.
We attribute it to the fact that the information between every two nodes in our proposed graph can be passed through in no more than two edges, and that increasing the size of randomly initialized parameters may not be beneficial for BERT fine-tuning.

To evaluate the effectiveness of our proposed model, we conduct an ablation study on the development dataset on the BERT-base setting. The results are shown in Table~\ref{table:nq-ablation}. First, we discuss the effect of the joint training strategy. We can see that the removal of either sub-task goals will bring decreases on both tasks. It suggests that the two-grained answers can promote each other with our multi-granularity representation.
Then we remove the whole graph module, which means the inference process depends on the score of short answer spans because long answer candidates cannot be scored. We can see the decrease of both long and short answer performance by $5.0\%$ and $0.9\%$, respectively, indicating the effectiveness of our proposed graph representations.

Finally, we investigate the effect of components in our graph encoder. In Table~\ref{table:nq-ablation}, we can see that without relational embedding, the performance on the long answer and short answer both slightly decrease. When removing the graph integration layer, the performance of long answer and short answer both become worse by $0.6\%$ and $0.8\%$. At last, we remove the self-attention layer in the graph encoder, the performance of long answer and short answer both become worse by $0.5\%$ and $0.7\%$. The ablation study shows the importance of each component in our method.


\section{Related Work}
Machine reading comprehension has been widely investigated 
since the release of large-scale datasets
\citep{rajpurkar2016squad,joshi2017triviaqa,lai2017race,trischler2017newsqa,yang2018hotpotqa}.
Lots of work has begun to build end-to-end deep learning models and has achieved good results
\citep{DBLP:conf/iclr/SeoKFH17,xiong2016dynamic,cui2017attention,devlin2018bert, lv_aaai20}.
They normally treat questions and documents as two simple sequences regardless of their structures and focus on incorporating questions into the documents, where the attention mechanism is most widely used.
\citet{DBLP:conf/acl/GardnerC18} proposes a model for multi-paragraph reading comprehension using TF-IDF as the paragraph selection method.
\citet{wang2018multi} focuses on modeling a passage at word and sentence level through hierarchical attention.

Previous work on document modeling is mainly based on a two-level hierarchy
\citep{ruder2016hierarchical,tang2015document,yang2016hierarchical,cheng2016neural,koshorek2018text,zhang2019pretraining}.
The first level encodes words or sentences to get the low-level representations. 
Moreover, a high-level encoder is applied to obtain document representation from the low-level.
In these frameworks, information flows only from low-level to high-level.
\citet{fernandes2018structured} proposed a graph neural network model for summarization and this framework allows much complex information flows between nodes, which represents words, sentences, and entities in the graph.  

Graph neural networks have shown their flexibility in a variant of NLP tasks \citep{DBLP:conf/emnlp/Zhang0M18,DBLP:conf/naacl/MarcheggianiBT18,DBLP:conf/acl/ZhangLS18,DBLP:conf/emnlp/SongZWG18}. 
A recent approach that began with Graph Attention Networks \citep{DBLP:conf/iclr/VelickovicCCRLB18}, which applies attention mechanisms to graphs. \citet{kgat_kdd19} proposed knowledge graph attention networks to model the information in the knowledge graph, \citep{zhang2018gaan} proposed gated attention networks, which use a convolutional sub-network to control each attention head's importance. 
We model the hierarchical nature of documents by representing them at four different levels of granularity. 
Besides, the relations between nodes are represented by different types of edges in the graph.
\section{Conclusion}

In this work, we present a novel multi-grained MRC framework based on graph attention networks and BERT. We model documents at different levels of granularity to learn the hierarchical nature of the document.
On the Natural Questions dataset, which contains two sub-tasks predicting a paragraph-level long answer and a token-level short answer, our method jointly trains the two sub-tasks to consider the dependencies of the two-grained answers.
The experiments show that our proposed methods are effective and outperform the previously existing methods by a large margin. 
Improving our graph structure of representing the document as well as the document-level pretraining tasks is our future research goals. Besides, the currently existing methods actually cannot process a long document without truncating or slicing it into fragments. How to model long documents is still a problem that needs to be solved.

\section*{Acknowledgments}

This work was supported by the National Natural Science Foundation of China (NSFC) via grant 61976072, 61632011 and 61772153.

\bibliography{manuscript}
\bibliographystyle{acl_natbib}

\vspace{0.1em}
\label{sec:appendix}
\appendix
\section*{Appendix}

\begin{table*}[h]
\centering
\small
\begin{tabular}{lccccccccccc}
\toprule
\multicolumn{1}{l}{} & \multicolumn{5}{c}{Long Answer} & \multicolumn{5}{c}{Short Answer} \\
 \multicolumn{1}{l}{} & Case1  & Case2  & Case3 & Case4  & Case5 & & Case1   & Case2   & Case3 & Case4  & Case5 \\ \midrule
BERT-base+Model-I   & 38.2 & 28.4 & 9.7 & 10.9 & 12.8 & &20.2  & 48.5  & \textbf{7.7} & 16.2 & 7.3 \\
BERT-base+Model-II & 40.8 & \textbf{28.6} & 8.4 & 9.7 & \textbf{12.5} & & 20.0 & \textbf{49.0} & \textbf{7.7} & 16.4 & \textbf{6.9} \\
BERT-base+Model-III         & \textbf{41.8} & \textbf{28.6} & \textbf{8.1} & \textbf{9.0}  & 12.6 & & \textbf{20.9}  & 48.2  & 8.0 & \textbf{15.3} & 7.7 \\ \midrule
BERT-syn+Model-I    & 40.0 & 30.0 & 7.9 & 11.0 & 11.1 & & 22.6  & \textbf{49.3} & \textbf{7.4} & 14.1 & \textbf{6.6} \\
BERT-syn+Model-II    & 42.8 & 30.7 & 6.6 & \textbf{9.5} & 10.4 & & 23.3  & 48.9 & 7.6 & 13.2 & 7.0 \\
BERT-syn+Model-III          & \textbf{43.0} & \textbf{30.9} & \textbf{6.2} & 9.7  & \textbf{10.2} & & \textbf{23.9}  & 48.2  & 8.1 & \textbf{12.2} & 7.7 \\
\bottomrule
\end{tabular}
\caption{Percentage of five categories for both long answer and short answer.}
\label{table:nq-error}
\end{table*}

\begin{table*}[h]
\centering
\small
\begin{tabular}{p{7.4cm}p{7.4cm}}
\toprule
\multicolumn{2}{l}{Question: what 's the dog 's name on tom and jerry} \\
\midrule
\textbf{Long Answer}: Tom ( named `` \textcolor{red}{Jasper} '' in his debut appearance ) is a grey and white domestic shorthair cat . `` Tom '' is a generic name for a male cat . He is usually but not always , portrayed as living a comfortable , or even pampered life , while Jerry ...
& 
\textbf{Long Answer:} \textcolor{blue}{Spike , occasionally referred to as Butch or Killer} , is a stern but occasionally dumb American bulldog who is particularly disapproving of cats , but a softie when it comes to mice ( though in his debut appearance , Dog Trouble , Spike goes after both Tom and Jerry ) ... \\
\textbf{Short Answer:} Jasper & \textbf{Short Answer:} Spike , occasionally referred to as Butch or Killer \\
\midrule
\multicolumn{2}{l}{Question: when is a spearman correlation meant to be used instead of a pearson correlation}\\
\midrule
\textbf{Long Answer:} This method should also not be used in cases \textcolor{red}{where the data set is truncated} ; that is , when the Spearman correlation coefficient is desired for the top X records ( whether by pre-change rank or post-change rank , or both ) , the user should use the Pearson correlation coefficient formula given above . 
& 
\textbf{Long Answer:}  The Spearman correlation between two variables is equal to the Pearson correlation between the rank values of those two variables ; while Pearson 's correlation assesses linear relationships , Spearman 's correlation \textcolor{blue}{assesses monotonic relationships ( whether linear or not )} ... \\
\textbf{Short Answer:} where the data set is truncated & \textbf{Short Answer:} assesses monotonic relationships ( whether linear or not ) \\
\bottomrule
\end{tabular}
\caption{Case studies from the development dataset. The results of directly predicting short answer span are shown on the left, and the results on the right are predicted by a pipeline strategy.}
\label{table:nq-case}
\end{table*}

\section{Error Analysis}

We provide an error analysis for our proposed models. We divide the results for instances in development dataset into five cases: 
\begin{itemize}
    \item Case 1: The question has a long (short) answer, and the predicted score is above the threshold.
    \item Case 2: The question does not have a long (short) answer, and the predicted score is below the threshold.
    \item Case 3: The question has a long (short) answer, and prediction is wrong.
    \item Case 4: The question has a long (short) answer, and the predicted score is below the threshold.
    \item Case 5: The question does not have a long (short) answer, and the predicted score is above the threshold.
\end{itemize}

The analysis results are shown in Table~\ref{table:nq-error}. For BERT-base+Model-III, we can see it outperforms other BERT-base models in the first four cases on the long answer and gets comparable results on Case 5. For the short answer, the improvement of our proposed model mainly comes from Case 1 and Case 4, which suggests that our approach helps the model do well in cases that have a short answer. Comparing Model-I and Model-III, we can see the significant improvement of our model lies in the long answer on Case 1 (From $38.2\%$, $40.0\%$ to $41.8\%$, $43.0\%$, respectively).

For Case 2 and Case 5, our Model-III does not have significant improvement compared to Model-I. The reason is that, for instances with no answer or no apparent answers, fine-grained information is more crucial.
Therefore, using the score of short answer spans might be more accurate than the long answer score from paragraph nodes, which are coarse-grained. Overall, our Model-III is better than the baseline Model-I, especially for examples with long or short answers.

\section{Case Study}

We report two case studies on the development dataset shown in Table~\ref{table:nq-case}. 
In the first case, the former prediction finds a wrong short answer ``Jasper'' where the word-level information in question ``name'' and ``tom'' is captured within a minimal context. 
Our pipeline strategy can consider the context of the whole paragraph, leading to a more accurate long answer along with its short answer. 
For the second case, the former prediction failed to capture the turning information while our pipeline model sees the whole context in the paragraph, which leads to the correct short answer. 
In both two cases, short answers on the left both have a larger score than those on the right.
This suggests that for a model that learns a strong paragraph-level representation, we can prevent errors from short answers by constraining it to the selected long answer spans.

\end{document}